\documentclass[10pt,twocolumn,letterpaper]{article}

\usepackage{cvpr}
\usepackage{times}
\usepackage{epsfig}
\usepackage{graphicx}
\usepackage{amsmath}
\usepackage{amssymb}

\usepackage{url}
\usepackage{booktabs}
\usepackage{enumerate}
\usepackage{color}
\usepackage{jeffe} 
\usepackage{subfigure}
\usepackage{multirow}
\usepackage{footnote} 

\usepackage[pagebackref=true,breaklinks=true,letterpaper=true,colorlinks,bookmarks=false]{hyperref}
\usepackage{cleveref}

\renewcommand{\baselinestretch}{0.983}

\newcommand{\vb}[1]{\mathbf{#1}}
\newcommand{\mc}[1]{\mathcal{#1}}

\newcommand{\WIP}[1]{}

\newcommand{\Ftr}{F_{tr}}
\newcommand{\Fvl}{F_{vl}}
\newcommand{\Fts}{F_{ts}}

\newcommand{\Nts}{U_{ts}}
\newcommand{\p}{P}
\newcommand{\xrimg}{\vb x} 
\newcommand{\yrcls}{y} 
\newcommand{\ximg}{\vb x} 
\newcommand{\ycls}{y} 
\newcommand{\xrimgF}{\xrimg_{\mc F}}
\newcommand{\xrimgN}{\xrimg_{\mc U}}

\newcommand{\ximgF}{\ximg_{\mc F}}

\newcommand{\yrclsF}{\yrcls_{\mc F}}
\newcommand{\yrclsN}{\yrcls_{\mc U}}
\newcommand{\yclsF}{\ycls_{\mc F}}

\makeatletter
\newcommand*{\storecounter}[2]{%
  \edef\@currentlabel{\the\value{#1}}
  \label{#2}
}
\makeatother

\cvprfinalcopy 

\def\cvprPaperID{7765} 

\ifcvprfinal\fi
\begin{document}

\title{Improving Confidence Estimates for Unfamiliar Examples}

\author{Zhizhong Li, Derek Hoiem \\
Department of Computer Science,\\
University of Illinois Urbana Champaign\\
{\tt\small \{zli115,dhoiem\}@illinois.edu}
}

\renewcommand{\baselinestretch}{0.99}
\setlength{\tabcolsep}{5pt}

\maketitle
\thispagestyle{empty}

\begin{abstract}
Intuitively, unfamiliarity should lead to lack of confidence. In reality, current algorithms often make highly confident yet wrong predictions when faced with relevant but unfamiliar examples.  A classifier we trained to recognize gender is 12 times more likely to be wrong with a 99\% confident prediction if presented with a subject from a different age group than those seen during training.  In this paper, we compare and evaluate several methods to improve confidence estimates for unfamiliar and familiar samples.  We propose a testing methodology of splitting unfamiliar and familiar samples by attribute (age, breed, subcategory) or sampling (similar datasets collected by different people at different times).  We evaluate methods including confidence calibration, ensembles, distillation, and a Bayesian model and use several metrics to analyze label, likelihood, and calibration error.  While all methods reduce over-confident errors, the ensemble of calibrated models performs best overall, and T-scaling performs best among the approaches with fastest inference. Our code is available at \url{https://github.com/lizhitwo/ConfidenceEstimates}.
\end{abstract}

\section{Introduction}
\label{sec:intro}

\begin{figure}[t]
  \centering
    \includegraphics[trim=0 0 0 0.25in,clip, width=0.15\textwidth]{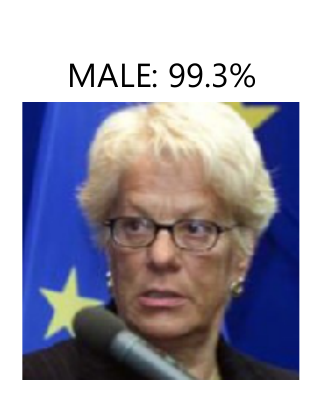}
    \includegraphics[trim=0 0 0 0.25in,clip, width=0.15\textwidth]{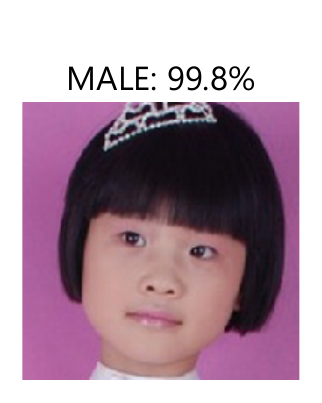}
    \includegraphics[trim=0 0 0 0.25in,clip, width=0.15\textwidth]{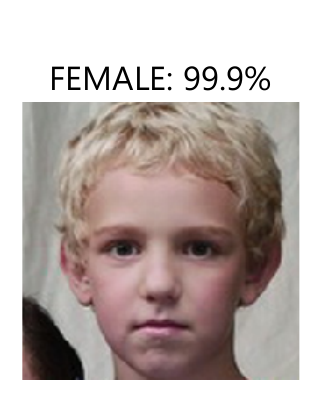}
    \includegraphics[trim=0 0 0 0.25in,clip, width=0.15\textwidth]{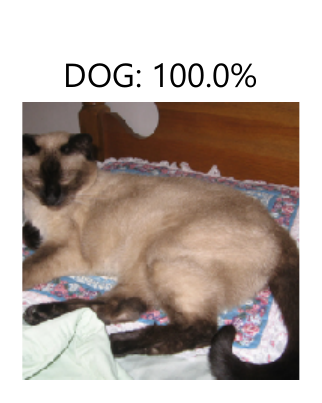}
    \includegraphics[trim=0 0 0 0.25in,clip, width=0.15\textwidth]{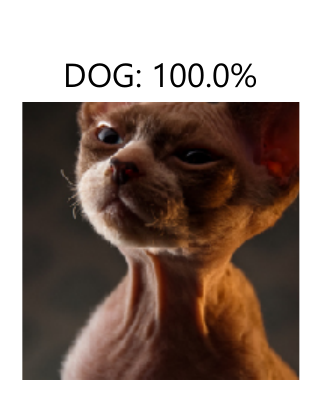}
    \includegraphics[trim=0 0 0 0.25in,clip, width=0.15\textwidth]{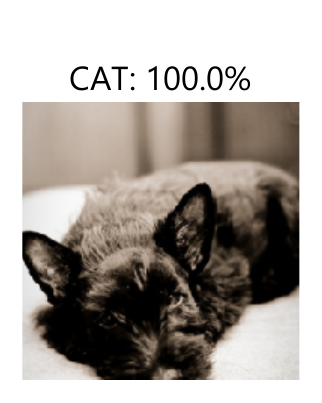}
    \includegraphics[trim=0 0 0 0.25in,clip, width=0.15\textwidth]{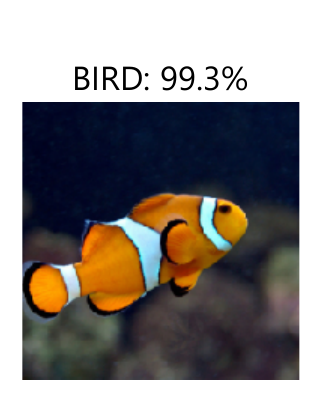}
    \includegraphics[trim=0 0 0 0.25in,clip, width=0.15\textwidth]{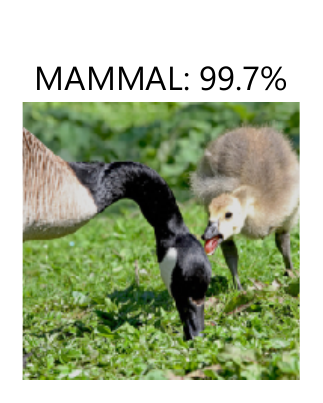}
    \includegraphics[trim=0 0 0 0.25in,clip, width=0.15\textwidth]{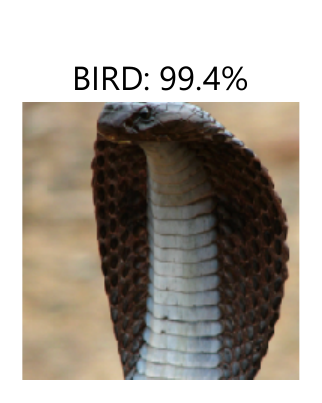}
\vspace{-0.12in}
  \caption{Deep networks often make highly confident mistakes when samples are drawn from outside the distribution observed during training. Examples shown have ages (top), breeds (middle), or species (bottom) that are not observed during training and are misclassified by a deep network model with high confidence. This paper investigates the problem of overconfidence for unfamiliar samples and evaluates several potential methods for improving reliability of prediction confidences.}
  \label{fig:intro}
  \vspace{-0.2in}
\end{figure}

In research, the i.i.d. assumption, that train and test sets are sampled from the same distribution, is convenient and easily satisfied. In practice, the training and test samples often come from different distributions, as developers often have access to a less diverse set of images than future samples observed by the deployed system. For example, the face images gathered by a company's employees  may not have the racial or age diversity of the world's population. Scholars that study the impact of AI on society consider differently distributed samples to be a major risk~\cite{varshney2017safety}:  ``This is one form of epistemic uncertainty that is quite relevant to safety because training on a dataset from a different distribution can cause much harm.'' Indeed, high profile failures, such as a person being labeled as a gorilla~\cite{newsgorilla} or a car driving through a tractor trailer~\cite{newstesla}, are due at least in part to failure to provide good confidence estimates for unfamiliar data.  

In this paper, our goal is to compare and evaluate several methods for improving confidence estimates for familiar and unfamiliar samples.  We consider {\em familiar samples} to be drawn from the same distribution as the training, as is typically done when creating training and test sets for research. We term {\em unfamiliar samples} as drawn from a different but still applicable distribution.  For example, for cat vs. dog classification, an image of a dog from a breed seen during training is familiar, while an image from a breed not seen during training is unfamiliar. We are not concerned with non-applicable ``out of domain'' images such as an image of a pizza for cat vs. dog classification.  

We propose familiar/unfamiliar splits for four image classification datasets and evaluate by measuring accuracy of predicted labels and confidences.  One would expect that classifiers would be less accurate and less confident for unfamiliar samples.  Our experiments confirm that deep network classifiers have lower prediction accuracy on unfamiliar samples but also show that wildly confident wrong predictions occur much more often, due to higher calibration error.  A simple explanation is that classifiers minimize a loss based on $P(y | x)$, for label $y$ and input features $x$, which is unregulated and unstable wherever $P(x)\sim 0$ in training. Empirical support for this explanation comes from Novak et al.~\cite{novak2018sensitivity} who show that neural networks are more robust to perturbations of inputs near the manifold of the training data.  We examine the effectiveness of calibration (we use temperature scaling~\cite{guo2017calibration}) for improving confidence estimates and the potential for further improvement using uncertainty-sensitive training~\cite{kendall2017uncertainties}, ensembles, and scaling based on novelty scores.  Since calibrated ensembles perform best but are most computationally expensive, we also investigate distilling the ensemble from a mix of supervised and unsupervised data.

Our paper's \textbf{key contributions} are: (1) highlight the problem of overconfident errors in practical settings where test data may be sampled differently than training; (2) propose a methodology to evaluate performance on unfamiliar and familiar samples; (3) demonstrate the importance of confidence calibration and compare several approaches to improve confidence predictions, including new ideas for incorporating novelty prediction and mixed supervision distillation.

\section{Related Work}
\label{sec:related}

\textbf{Unreliability of prediction for unfamiliar samples: } Lakshminarayanan et al.~\cite{lakshminarayanan2017simple} show that networks are unreliable when tested on semantically unrelated or out-of-domain samples, such as applying object classification to images of digits. They also show that a using the Brier score~\cite{Brier1950VERIFICATIONOF} (squared error of 1 minus confidence in true label) as a loss and training an ensemble of classifier improves confidence calibration and reduces overconfident errors on out-of-domain samples. Ovadia et al.~\cite{ovadia2019}, in independent work concurrent to ours, also find that ensembles are most effective for skewed and out-of-domain samples, evaluating with Brier score, negative log likelihood of predictions, and expected calibration error (ECE). Our inclusion of Brier score and ECE is inspired by these methods. Our paper differs from these in the consideration of natural (not artificially distorted) samples from unfamiliar but semantically valid distributions, which is a common practical scenario when, for example, developers and users have access to different data.  Roos et al.~\cite{roos2006generalization} distinguish between i.i.d. generalization error and off-training-set error and provide bounds based on repetition of input features.  Extending their analysis to high dimensional continuous features is a worthwhile area of further study. 

\textbf{Methods to address epistemic uncertainty: } When faced with unfamiliar samples, a model suffers from 
{\em epistemic uncertainty}, the uncertainty due to incomplete knowledge. Related works reduce this uncertainty by averaging over several models, with the intuition that different models will disagree and thus appropriately reduce certainty for parts of the feature distribution that are not well represented by the training set.  Bayesian approaches~\cite{bishop1995neural,blundell2015weight,hernandez2016black} estimate a distribution of network parameters and produce a Bayesian estimate for likelihood. These methods are usually very computationally intensive~\cite{lakshminarayanan2017simple}, limiting their practical application. Gal and Ghahramani~\cite{gal2016dropout} propose MC-dropout as a discrete approximation of Bayesian networks, using dropout for a Monte-Carlo sample of likelihood estimates. Follow-up work by Kendall and Gal~\cite{kendall2017uncertainties} proposes to estimate both aleatoric and epistemic uncertainties to increase the performance and quality of uncertainty. Lakshminarayanan et al.~\cite{lakshminarayanan2017simple} propose a simpler method to average over predictions from multiple models with an ensemble of deep networks, an approach further validated by Ovadia et al.~\cite{ovadia2019}.  Multi-head deep networks~\cite{lee2015m,makansi2019overcoming} emulate ensembles and are shown to outperform MC-dropout. Hafner et al.~\cite{hafner2018reliable} propose a loss that encourages high uncertainty on training samples whose features are permuted or perturbed by noise. Our work differs primarily in its investigation of unfamiliar samples that are differently distributed from training but still have one of the target labels. Concurrent work by Mukhoti et al.~\cite{mukhoti2019} proposes combining focal loss with T-scaling.  
We evaluate T-scaling calibration~\cite{guo2017calibration}, Kendall and Gal~\cite{kendall2017uncertainties}, ensembles, and a proposed novelty-sensitive T-scaling approach.

\textbf{Calibration} methods aim to improve confidence estimates, by learning a mapping from prediction scores to a well-calibrated probability. We use 
{\em T-scaling}, short for temperature scaling, in which the logit score of a classifier is divided by a scalar $T$ as a special case of Platt calibration~\cite{platt1999probabilistic}. In a broad evaluation of calibration methods, Guo et al.~\cite{guo2017calibration} found T-scaling to be the simplest and most effective.  Note that T-scaling has no effect on the rank-order of predictions, so it affects only the Brier error, negative log likelihood, and expected calibration error, not label error.  Calibration parameters are fit to the validation set which is i.i.d. with training.  Thus, calibration does not explicitly deal with unfamiliar samples, but our experiments show that calibration is an essential part of the solution for producing accurate confidence estimates on both familiar and unfamiliar samples.

\textbf{Distillation}~\cite{hinton2015distilling} regresses the confidence predictions of a network to match those of another model, such as a larger network or ensemble.  Radosavovic \etal~\cite{radosavovic2017data} obtain soft labels on transformed unlabeled data and use them to distill for unsupervised learning. Li and Hoiem~\cite{li2017learning} extend models to new tasks without retaining old task data, using the new-task examples as unsupervised examples for the old tasks with a distillation loss. Distillation has also been used to reduce sensitivity to adversarial examples that are similar to training examples~\cite{papernot2016distillation}. We investigate whether distilling an ensemble into a single model can preserve the benefits of the ensemble on familiar and unfamiliar data, when using the training set and an additional unsupervised dataset to distill.

\textbf{Other: } The remainder of this section describes works and problem domains that are less directly related. 
{\em Domain adaptation} (e.g.,~\cite{quionero2009dataset}) aims to train on a source domain and improve performance on a slightly different target domain, either through unsupervised data or a small amount of supervised data in the target domain. {\em Domain generalization}~\cite{muandet2013domain,li2017deeper,shankar2018generalizing} aims to build models that generalize well on a previously unspecified domain, whose distribution can be different from all training domains. These models generally build a domain-invariant feature space~\cite{muandet2013domain} or a domain-invariant model~\cite{shankar2018generalizing}, or factor models into domain-invariant and domain-specific parts~\cite{li2017deeper}. Attribute-based approaches, such as Farhadi \etal~\cite{farhadi2009describing}, attempt to learn features or attributes that are more likely to be consistent between familiar and unfamiliar samples. These methods require multiple training domains to learn invariant representations, with the intent to improve robustness to variations in the target domain.  {\em One-shot learning} (e.g.~\cite{vinyals2016matching}) and {\em zero-shot learning} (e.g.~\cite{Zendel_2017_CVPR}) aim to build a classifier through one sample or only metadata of the class. 
Many methods more broadly attempt to {\em improve generalization}, such as data augmentation or jittering~\cite{pomerleau1997}, dropout~\cite{srivastava2014dropout}, batch normalization~\cite{ioffe2015batchnorm}, and weight decay. Hoffer \etal~\cite{hoffer2017train} propose better hyperparameter selection strategies for better generalization. Bagging~\cite{Breiman1996}, ensembles, and other model averaging techniques are also used prior to deep learning. 

Other methods aim to reduce confident errors by detecting failure~\cite{fumera2002support,geifman2017selective,zhang2014predicting,wang2018towards,scheirer2011meta}, for example by looking at how close samples are to decision boundaries or estimating whether a test sample comes from the same distribution as training~\cite{devarakota2007confidence,lee2012object,liang2018enhancing,tax2001one,khan2009survey}.  Typically, the motivation of these methods is to avoid making any prediction on suspect samples, while the goal of our work is to understand and improve performance of classifier predictions on both familiar and unfamiliar samples that have applicable labels.

\section{Problem Setup and Methods}
\label{sec:method}

In many commercial settings, the developers of an algorithm have access to data that may be limited by geography, demographics, or challenges of sampling in diverse environments, while the intended users, in aggregate, have much broader access.  For example, developers of a face attribute classification algorithm may undersample children, elderly, or Inuits, due to their own demographics. Someone training a plant recognition algorithm may have difficulty collecting samples of species not locally native.  A recognizer of construction equipment may be applied to vehicle models that came out after release of the classification model.  To study and improve the robustness of classifiers in these settings we explore:
\begin{itemize}
    \itemsep0pt    
    \item How to organize data to simulate the familiar and unfamiliar test sets (Sec.~\ref{sec:data})
    \item How to evaluate the quality of predictions (Sec.~\ref{sec:metrics})
    \item What methods are good candidates to improve prediction quality on unfamiliar samples (Sec.~\ref{sec:methods})
\end{itemize}


\subsection{Datasets and Familiar/Unfamiliar Split}
\label{sec:data}

\begin{figure*}[t]
  \centering
    \includegraphics[width=0.8\textwidth]{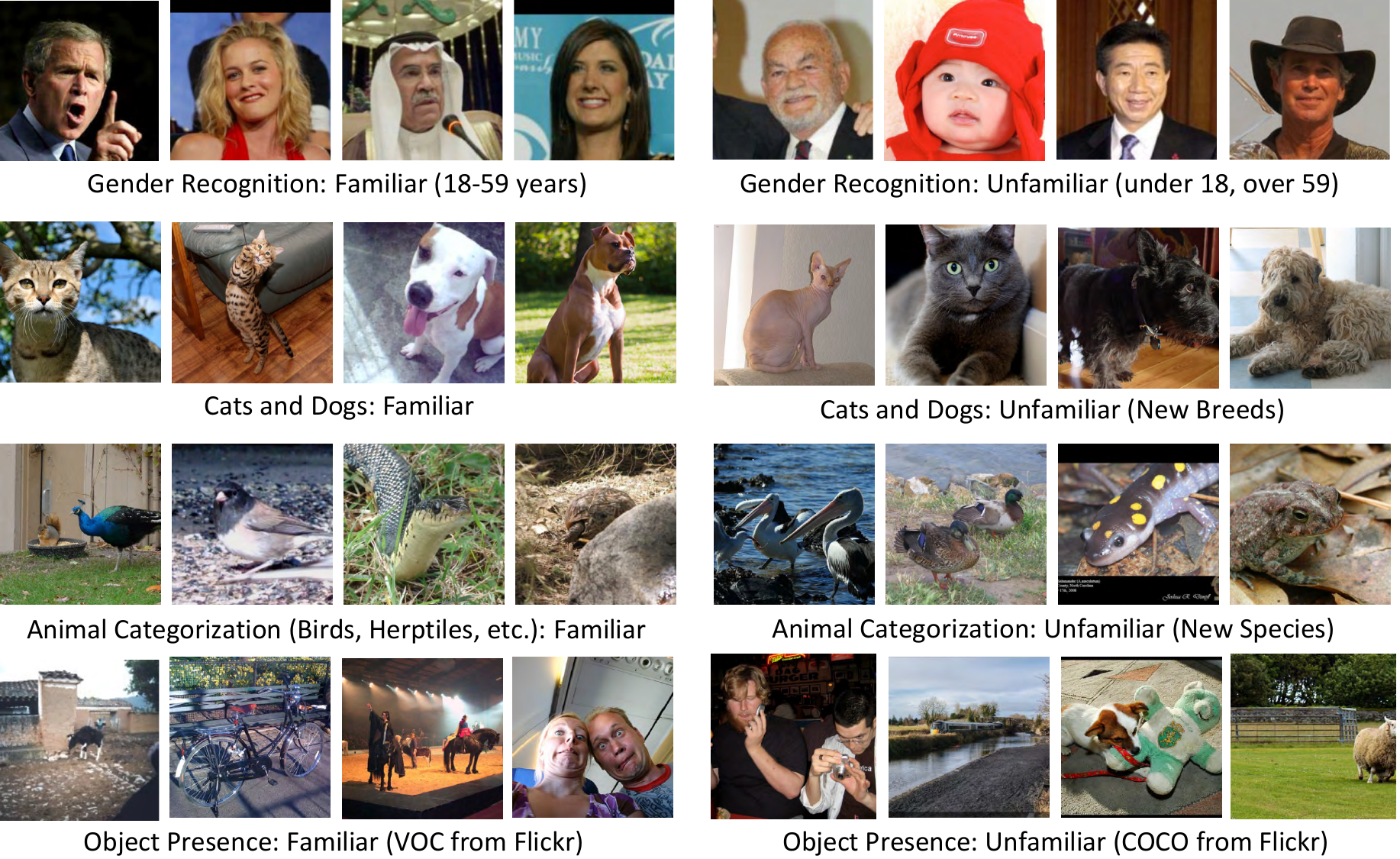}
  \caption{Familiar and unfamiliar samples from each dataset.  To study how classifier performance varies with novelty, we create splits of unfamiliar and familiar samples that are task-relevant, where the split is defined by age, breed, species, or date of sampling.  The first three represent cases where the training distribution does not fully cover the test cases.  The last represents a case of the minimal novelty achievable without independently sampling from the same image set.}
  \label{fig:dataset}
  \vspace{-0.1in}
\end{figure*}

We choose four classification tasks for evaluation, detailed below and shown in Figure~\ref{fig:dataset}.  For each of the first three tasks, a dataset is first split into ``familiar'' and ``unfamiliar'' subsets according to an attribute or subcategory, simulating the case of training data not containing the full diversity of potential inputs. In the fourth task (object presence classification), two similar datasets are used for the same object categories, simulating similar sources but sampled at different times. The ``familiar'' samples $(\xrimgF,\yrclsF) \sim \mc{F}$ are further split into training $\Ftr$, validation $\Fvl$, and test $\Fts$ sets, while the ``unfamiliar'' samples $(\xrimgN,\yrclsN) \sim \mc{U}$ are used only for testing. The inputs $\xrimgF$ and $\xrimgN$ may occupy different portions of the feature space, with little to no overlap, but where they do overlap $\p(\yrcls|\xrimg)$ is the same for $\mc F$ and $\mc U$. No sample from $\mc U$ is ever used in pre-training, training, or validation (parameter selection). In some cases, we use a dataset's standard validation set for testing (and not parameter tuning) so that we can compute additional metrics, as ground truth is not publicly available for some test sets.

\textbf{Gender} recognition:  The extended Labeled Faces in the Wild (LFW+) dataset~\cite{HanMSUTR14} with 15,699 faces is used. Samples are split into familiar $\mc{F}$ and unfamiliar $\mc{U}$ based on age annotations provided by Han et al.~\cite{HanTPAMI17}, with familiar ages 18-59 years and unfamiliar ages outside that range.  The dataset comes with five preset folds; we use the first two for training, the third fold for validation, and the last two for testing.

\textbf{Cat vs. dog} recognition: Using the Pets dataset~\cite{parkhi12a}, the first 20 dog breeds and first 9 cat breeds are familiar, and the other 5 dog and 3 cat breeds are unfamiliar. The standard train/test splits are used (with training samples from $\mc U$ excluded).

\textbf{Animal} categorization:  Four animal superclasses (mammals, birds, herptiles, and fishes) are derived from ImageNet~\cite{ILSVRC15}, and different subclasses are used for familiar and unfamiliar sets.  After sorting object classes within each superclass by their indices, the first half of classes are familiar $\mc F$, and the second half are unfamiliar $\mc U$.  The data is also subsampled, so there are 800 training and 200 validation examples drawn from the ImageNet training set per superclass, and 400 examples drawn from the ImageNet validation set for each of the unfamiliar and familiar test sets.  

\textbf{Object} presence classification:  The PASCAL VOC 2012 dataset~\cite{Everingham15} is used as familiar, with the similar 20 classes in MS COCO~\cite{lin2014microsoft} used as unfamiliar.  \verb|tvmonitor| is mapped to \verb|tv|. Test samples are drawn from the VOC PASCAL and MS COCO validation sets.  The familiar and unfamiliar samples in this task are more similar to each other since they vary, not by attribute or subclass, but by when and by whom the images were collected.

\subsection{Evaluation Metrics}
\label{sec:metrics}

We use several error metrics to assess the quality of classifier predictions. We denote $\p_m(\yrcls_i|\xrimg_i)$ as the assigned confidence in the correct label for the $i^{th}$ of $N$ samples by a model $m$.  In all metrics, lower is better.

\begin{figure}[t]
  \centering
    \includegraphics[width=0.8\columnwidth]{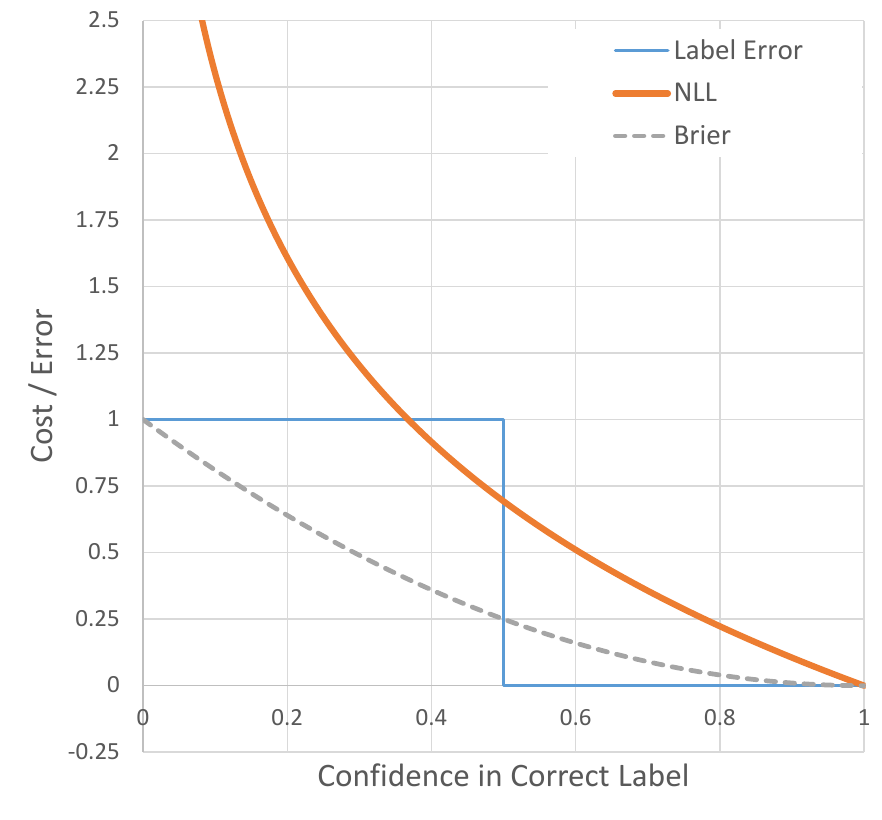}
    \vspace{-0.05in}
  \caption{\textbf{Prediction quality metrics}: plot of error vs. confidence in correct label for 0-1 classification. NLL (negative log likelihood) strongly penalizes confidently wrong predictions, while Brier error penalties are constrained. Label error does not assess confidence beyond which label is most likely.}
  \label{fig:metrics}
  \vspace{-0.2in}
\end{figure}

\textbf{NLL: } Negative log likelihood (NLL) $\frac{1}{N}\sum_{i} -\log \p_m(\yrcls_i|\xrimg_i)$ is a natural measure of prediction quality and commonly used as a loss for training classification models (often called ``cross-entropy''), as it corresponds to the joint probability of predictions on independently drawn samples. The main drawback is that NLL is unbounded as confidence in the correct class approaches 0.  To help remedy this, we clip the softmax probability estimates to $[0.001,0.999]$ for  all models before computing NLL. 

\textbf{Brier: } The Brier score~\cite{Brier1950VERIFICATIONOF} measures the root mean squared difference between one and the confidence in the correct label: $\left(\frac{1}{N}\sum_{i} (1 - \p_m(\yrcls_i|\xrimg_i))^2\right)^{1/2}$.  Similar to NLL, Brier is smallest when the correct label is predicted with high confidence, but the penalty for highly confident errors is bounded at 1, avoiding too much emphasis on a few large errors.  We use RMS (root mean squared) instead of mean squared, as in the original, because we find it easier to interpret, and we call it ``Brier error'', since it should be minimized. 

\textbf{Label Error: } Label error is measured as the percent of incorrect most likely labels, or 1 minus average precision. We use percent incorrect for all tasks except object presence classification, for which we use mean average precision, in accordance with community norms for reporting performance on these tasks.

\textbf{ECE: } Expected calibration error (ECE) measures whether the classifier ``knows what it knows''. Following the notation of~\cite{guo2017calibration}, ECE is computed as $\sum_{j=1}^J\frac{|B_j|}{N}\lvert\rm{acc} \mathit{(B_j)}-\rm{conf} \mathit{(B_j)}\rvert$ where $B_j$ is a set of predictions binned by confidence quantile, $\rm{acc}\mathit{(B_j)}$ is the average accuracy of the $B_j$, and $\rm{conf}\mathit{(B_j)}$ is the average confidence in the most likely label.  We use 10 quantiles for binning.

\textbf{E99: } E99 is the error rate among the subset of samples that have at least 99\% confidence in any label. If the classifier is well-calibrated, E99 should be less than 1\%. We created E99 to directly measure a model's tendency to generate highly confident errors.

\subsection{Compared Methods}
\label{sec:methods}

Deep network classifiers are often overconfident, even on familiar samples~\cite{guo2017calibration}. On unfamiliar samples, the predictions are less accurate and even more overconfident, as our experiments show. We consider several tools to improve predictions: calibration, novelty detection, ensembles, and loss functions that account for uncertainty. Some methods provide better confidence calibration, while others (e.g. ensembles and Bayesian models) can also provide more confidently accurate predictions. 

\textbf{T-scaling: } Calibration aims to improve confidence estimates so that a classifier's expected accuracy matches its confidence.  Among these, we use the temperature-scaling method described in Guo et al.~\cite{guo2017calibration}. At test time, all softmax logits are divided by temperature $T$.  With $T>1$, prediction confidence is decreased. $T$ is a single parameter set to minimize NLL on the validation set.  We then use this $T$ on a network retrained on both training and validation sets.  

\textbf{Novelty-weighted scaling: } We also consider novelty-weighted scaling, with the intuition that confidence should be lower for novel (i.e. unfamiliar) samples than for those well represented in training. We use the ODIN~\cite{liang2018enhancing} model-free novelty detector.  Since the novelty scores $novelty(\ximg)$ often have a small range, we normalize them by linearly scaling the $5^{th}$ and $95^{th}$ percentile on training data to be 0 and 1 and clipping values outside $[0, 1]$. We then modify temperature scaling to set $T(\ximg) = T_0 + T_1 \cdot  novelty(\ximg)$, with $T_0$ and $T_1$ set by grid search on the validation set, so that temperature depends on novelty. 

\textbf{Ensemble} methods consider both model parameter and data uncertainty by averaging over predictions.  In areas of the feature space that are not well represented by training data, members of the ensemble may vary in their predictions, reducing confidence appropriately.  In our experiments, members of the ensemble are trained with all training samples and differ due to varying initialization and stochastic optimization.  We found this simple averaging approach to outperform bagging and bootstrapping. In prediction, the member confidences in a label $y_i$ are averaged to yield the ensemble confidence: $\p_m(y_i|\xrimg_i)=\frac{1}{M}\sum_j^M\p_{m_j}(y_i|\xrimg_i)$, where $M$ is the number of ensembles. $M=10$ in our experiments. 

\textbf{Distillation: } Our experiments show the ensemble is highly effective, but it is also $M$ times more expensive for inference.  We, thus, consider whether we can retain most of the benefit of the ensemble at lower compute cost using distillation~\cite{hinton2015distilling}.  After training the ensemble, the distilled model is trained by minimizing a weighted distillation loss (minimizing temperature-scaled cross-entropy of the ensemble's soft predictions with the distilled model's predictions) and a classification loss:
\begin{align}
\begin{split}
     \mc{L} =  \dfrac{1}{|\Ftr|}\sum_{(\ximgF,\yclsF) \in \Ftr} \bigg( & \lambda_{cls} \mc{L}_{cls}(\yclsF, f_{dis}(\ximgF)) +  \\  &\mc{L}_{dis}(f_{ens}(\ximgF), f_{dis}(\ximgF)) \bigg)
\end{split}
\end{align}

where $\mc{L}_{cls}$ is the classification loss over the distilled model's soft predictions $f_{dis}(\ximgF)$, $\mc{L}_{dis}$ is the distillation loss over the soft predictions of the distilled model and ensemble $f_{dis}(\ximgF)$, and $\lambda_{cls}$ is a weighting to balance classification and distillation losses ($\lambda_{cls}$ = 0.5, as recommended in~\cite{hinton2015distilling}).

\textbf{G-distillation: } Under the standard distillation, the distilled model is guided to make similar predictions to the ensemble for the familiar distribution $\mc F$, but its predictions are still unconstrained for unfamiliar samples, potentially losing the benefit of the ensemble's averaging for samples from $\mc U$.  Therefore, we propose G-distillation, a generalized distillation where the distillation loss is also computed over samples from an unsupervised distribution $\mc G$. In our experiments, we choose $\mc G$ to be related to the task, but make sure there is no overlap between specific examples in $\mc G$ and $\Fts$ or $\mc U$.  We use the following unsupervised datasets for $\mc G$ in our experiments: Gender, CelebA~\cite{liu2015faceattributes}; Broad animal, COCO~\cite{lin2014microsoft}; Cat-dog, ILSVRC12~\cite{ILSVRC15}; and Object presence, Places365-standard~\cite{zhou2017places}. The images from $\mc G$ are disjoint with the datasets used to draw $\mc F$ and $\mc U$ for each respective task.

\textbf{Bayesian model: } Finally, we consider the Bayesian method of Kendall et al.~\cite{kendall2017uncertainties}, which accounts for uncertainty in model parameters (epistemic) and observations (aleatoric).  To account for model parameter uncertainty, multiple predictions are made with Monte Carlo Dropout, and predictions are averaged.  In this way, dropout is used to simulate an ensemble within a single network.  In our implementation, we apply dropout to the second-to-last network layer with a rate of 0.2.  Observation uncertainty is modeled with a training loss that includes a prediction of logit variance. The logits can then be sampled based on both dropout and logit variance, and samples are averaged to produce the final confidence. See \cite{kendall2017uncertainties} for details.

\section{Experiments}

When comparing these methods, we aim to answer the following experimental questions:
\begin{itemize}
    \itemsep0em
    \item Do T-scaling calibration parameters learned from $\mc F$ also improve confidence estimates on $\mc U$?   
    \item Does novelty-weighted scaling outperform the data agnostic T-scaling?
    \item Do ensembles learned on $\mc F$ also improve predictions on $\mc U$?
    \item Is distillation able to preserve ensemble performance on $\mc F$ and $\mc U$?
    \item Does adding the unsupervised set for distillation in G-distillation lead to better preservation?
    \item Does the Bayesian model that is specifically designed to manage model and observational uncertainty outperform more general alternatives?
\end{itemize}

\noindent (Spoiler alert: answers in order are yes, no, yes, no, yes, partially.)

\subsection{Training and Testing Details}

\textbf{Training: } For all experiments we use PyTorch with a ResNet-18 architecture and stochastic gradient descent\footnote{Errata: previous versions (v4, v5) incorrectly stated that Adam was used. In fact, we used SGD.} optimization with a momentum of $0.9$. We initialize the final layer of our pre-trained network using Glorot initialization~\cite{glorot2010understanding}. We perform hyper-parameter tuning for the learning rate and the number of epochs using a manual search on a validation split of the training data.\footnote{Errata: a previous version (v3) incorrectly included unfamiliar data in the T-scaling validation. Please disregard that result.}  When the performance on validation plateaus, we reduce the learning rate by a factor of 10 and run 1/3 as many additional epochs as completed up to that point.  After fitting hyperparameters on the validation data, the models are retrained using both train and val sets.  For data augmentation, we use a random crop and mirroring similar to Inception~\cite{szegedygoing}. 
Places365-standard~\cite{zhou2017places} dataset is used to pretrain the network, and the network is fine tuned separately for each task.  When training G-distillation, we sample the image from $G$ to be roughly $\frac 1 4$ the size of $F_{tr}$.  We also verified that using a different architecture (DenseNet161~\cite{huang2017densely}) yields the same  experimental conclusions. 

\textbf{Testing: } At test time we evaluate on the center crop of the image. Due to the relatively high variance of NLL on $\Nts$, we run our experiments 10 times to ensure statistical significance (unpaired two-tail t-test with p=0.95 on model performance), but we run the ensemble method only once (variance estimated using ensemble member performance variance). Our 10 runs of the distillation methods use the same ensemble run.

\subsection{Results}

Our main table of results is shown in Table~\ref{tab:bigtable}.  The \textbf{baseline} is a single uncalibrated ResNet-18 network. The others correspond to the methods described in Sec.~\ref{sec:method}. For the baseline, we show the absolute error, and for the other methods, we show the percent reduction in error compared to the baseline (e.g. a drop from 0.10 to 0.09 is a 10\% reduction) to facilitate comparison. The complete table with absolute error is included in the supplemental material. All methods except baseline use calibration. 
\begin{table}[t]
\footnotesize
\setlength{\tabcolsep}{3.5pt}
\begin{tabular}{lrrrrrrrr}
 & \multicolumn{2}{c}{\textbf{NLL}} & \multicolumn{2}{c}{\textbf{Brier}} & \multicolumn{2}{c}{\textbf{Label Error}} & \multicolumn{2}{c}{\textbf{ECE}} \\
\textbf{Gender} & \textbf{fam.} & \textbf{unf.} & \textbf{fam.} & \textbf{unf.} & \textbf{fam.} & \textbf{unf.} & \textbf{fam.} & \textbf{unf.} \\
Baseline & 0.083 & 0.542 & 0.147 & 0.352 & 0.028 & \textbf{0.147} & 0.013 & 0.109 \\
T-scaling & 12\% & 26\% & 2\% & \textbf{4\%} & 0\% & \textbf{0\%} & \textbf{73\%} & 20\% \\
Ensemble & \textbf{24\%} & 33\% & \textbf{10\%} & \textbf{6\%} & \textbf{22\%} & \textbf{0\%} & 36\% & \textbf{29\%} \\
Distill & 8\% & 33\% & 3\% & 4\% & 3\% & -7\% & 41\% & 21\% \\
G-distill & 13\% & \textbf{38\%} & 5\% & \textbf{6\%} & 9\% & -5\% & 31\% & \textbf{31\%} \\
Bayesian & 17\% & 26\% & 5\% & 4\% & 6\% & \textbf{0\%} & \textbf{77\%} & 19\% \\
 &  &  &  &  &  &  &  &  \\
\multicolumn{2}{l}{\textbf{Cat vs. Dog}} &  &  &  &  &  &  &  \\
Baseline & 0.053 & 0.423 & 0.112 & 0.290 & 0.016 & 0.095 & 0.010 & 0.078 \\
T-scaling & 23\% & 30\% & 4\% & 5\% & 0\% & 0\% & 64\% & 23\% \\
Ensemble & \textbf{40\%} & \textbf{46\%} & \textbf{17\%} & \textbf{12\%} & \textbf{22\%} & \textbf{8\%} & \textbf{79\%} & \textbf{46\%} \\
Distill & -13\% & 22\% & -9\% & 1\% & -18\% & -4\% & 55\% & 26\% \\
G-distill & -18\% & 27\% & -14\% & 1\% & -33\% & -8\% & 41\% & 31\% \\
Bayesian & 17\% & 26\% & 3\% & 5\% & 0\% & 3\% & 42\% & 21\% \\
 &  &  &  &  &  &  &  &  \\
\textbf{Animals} &  &  &  &  &  &  &  &  \\
Baseline & 0.326 & 1.128 & 0.199 & 0.341 & 0.104 & 0.291 & 0.048 & 0.187 \\
T-scaling & 13\% & 23\% & 3\% & 5\% & 0\% & 0\% & \textbf{75\%} & 37\% \\
Ensemble & \textbf{22\%} & \textbf{32\%} & \textbf{9\%} & \textbf{8\%} & \textbf{11\%} & \textbf{6\%} & 50\% & \textbf{57\%} \\
Distill & 7\% & 24\% & 1\% & 5\% & -1\% & 0\% & 66\% & 45\% \\
G-distill & 14\% & 26\% & 5\% & 7\% & 7\% & 2\% & 56\% & 49\% \\
Bayesian & 16\% & 24\% & 5\% & 5\% & 4\% & 1\% & \textbf{74\%} & 39\% \\
 &  &  &  &  &  &  &  &  \\
\textbf{Objects} &  &  &  &  &  &  &  &  \\
Baseline & 0.086 & 0.128 & 0.154 & 0.186 & 0.195 & 0.455 & 0.005 & 0.010 \\
T-scaling & 0\% & 0\% & 0\% & 0\% & 0\% & 0\% & 2\% & 2\% \\
Ensemble & \textbf{4\%} & 4\% & \textbf{2\%} & 2\% & \textbf{6\%} & \textbf{3\%} & \textbf{3\%} & 7\% \\
Distill & -1\% & \textbf{5\%} & 0\% & \textbf{2\%} & 1\% & 0\% & -31\% & \textbf{10\%} \\
G-distill & -2\% & 5\% & -1\% & 2\% & -2\% & -1\% & -41\% & 7\% \\
Bayesian & 0\% & 0\% & 0\% & 0\% & 0\% & 0\% & 3\% & 1\%
\end{tabular}
\hrule
\vspace{0.02in}
\caption{
\normalsize
Performance of baseline (single model) for several metrics and percent reduction in error for other methods. All methods except baseline use T-scaling calibration.  ``T-scaling'' is a single calibrated model.  
\label{tab:bigtable}
}
\end{table}

\begin{figure}[t]
  \centering
    \includegraphics[width=\columnwidth]{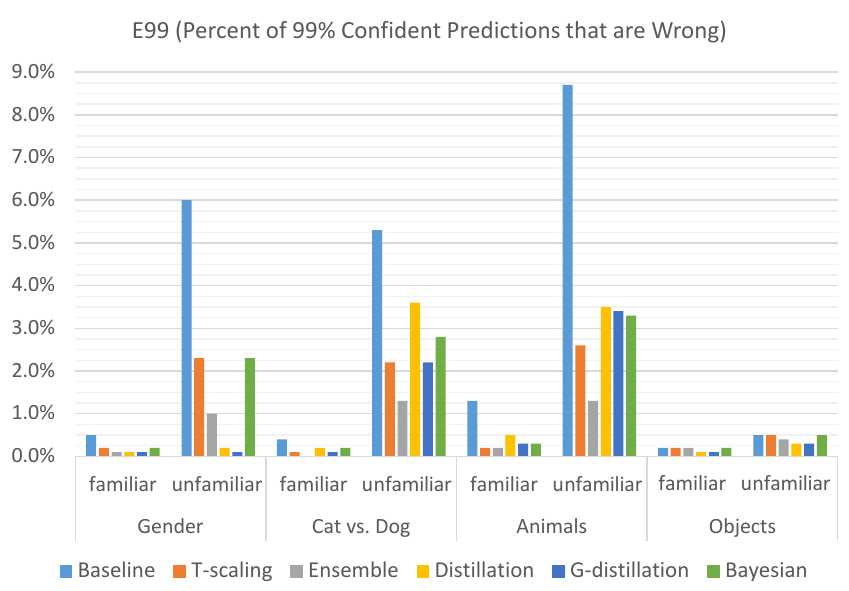}
  \caption{Classifiers are much more prone to confident errors when faced with unfamiliar samples. T-scaling calibration, among other methods, reduces the overconfidence, with ensembles of calibrated models providing consistent further improvement.  }
  \label{fig:e99}
  \vspace{-0.1in}
\end{figure}

\textbf{Familiar vs. Unfamiliar Performance:} Looking at baseline performance in Table~\ref{tab:bigtable}, we see much higher error rates  for unfamiliar samples, compared to familiar, for all tasks. The label error and calibration error are both higher, leading to much higher NLL and Brier error.  {\em This means the baseline classifier is less accurate and has poor ability to detect its own inaccuracy on unfamiliar samples} --- it does not know what it does not know.  For example, in gender recognition, the label error increases from 2.8\% for unfamiliar to 14.7\%; the calibration error ECE increases from 0.013 to 0.109; and the NLL increases from 0.083 to 0.542.  Figure~\ref{fig:e99} underscores the prevalence of confident errors, which are several times more common for unfamiliar samples than familiar. 

The differences between unfamiliar and familiar for object presence classification are substantial but smaller than other tasks, as expected, since VOC (familiar) and COCO (unfamiliar) images were both sampled from Flickr using similar methodologies~\cite{lin2014microsoft}.  The larger differences in mean AP (label error) may be due to lower frequency for a given object category in COCO.

\begin{table}[]
\small
\setlength{\tabcolsep}{5pt}
\begin{tabular}{lrrrrrr}
                     & \multicolumn{2}{c}{\textbf{NLL}} & \multicolumn{2}{c}{\textbf{Brier}} & \multicolumn{2}{c}{\textbf{ECE}} \\
\textbf{Gender}      & \textbf{fam.}  & \textbf{unf.}  & \textbf{fam.}   & \textbf{unf.}   & \textbf{fam.}  & \textbf{unf.}  \\
Single               & 0.083          & 0.542           & 0.148           & 0.352            & 0.013          & 0.109           \\
Sin. T-scale       & 0.073          & 0.400           & 0.145           & 0.338            & 0.004          & 0.087           \\
Ensemble             & 0.062          & 0.455           & 0.130           & 0.344            & 0.003          & 0.093           \\
Ens. T-scale     & 0.063          & 0.363           & 0.130           & 0.333            & 0.009          & 0.077           \\
                     &                &                 &                 &                  &                &                 \\
\textbf{Cat vs. Dog} &                &                 &                 &                  &                &                 \\
Single               & 0.053          & 0.423           & 0.110           & 0.290            & 0.010          & 0.078           \\
Sin. T-scale       & 0.041          & 0.295           & 0.105           & 0.276            & 0.004          & 0.060           \\
Ensemble             & 0.033          & 0.286           & 0.095           & 0.263            & 0.002          & 0.055           \\
Ens. T-scale     & 0.032          & 0.229           & 0.095           & 0.255            & 0.002          & 0.042           \\
                     &                &                 &                 &                  &                &                 \\
\textbf{Animals}     &                &                 &                 &                  &                &                 \\
Single               & 0.326          & 1.128           & 0.200           & 0.341            & 0.048          & 0.187           \\
Single T-scale       & 0.284          & 0.866           & 0.195           & 0.324            & 0.012          & 0.118           \\
Ensemble             & 0.256          & 0.930           & 0.182           & 0.322            & 0.022          & 0.138           \\
Ens. T-scale     & 0.254          & 0.772           & 0.182           & 0.311            & 0.024          & 0.080           \\
\end{tabular}
\hrule
\vspace{0.02in}
\caption{
\normalsize
\label{tab:calibration}
T-scaling calibration effectively reduces likelihood error (NLL, Brier) and calibration error (ECE) for many models across tasks for familiar and unfamiliar samples.  Without calibration, using an ensemble reduces these errors, but an ensemble of calibrated models (``Ens. T-scale'') performs best.  Applying T-scaling to an ensemble of uncalibrated classifiers, and creating an ensemble of calibrated classifiers produces nearly identical results.
}
\vspace{-0.2in}
\end{table}

\textbf{Importance of calibration: } Table~\ref{tab:calibration} compares performance of the baseline and ensemble methods, both without and with T-scale calibration.  Calibrated single models outperform uncalibrated models, and ensembles of calibrated models outperform ensembles of uncalibrated models. For example, in cat vs. dog recognition, the baseline NLL drops from 0.423 to 0.295, a 30\% reduction; and the ensemble NLL drops from 0.286 to 0.229, a 20\% reduction. Though not shown, a calibrated ensemble of uncalibrated models performs very similarly to an ensemble of calibrated models.   For the object presence task, there is little effect of calibration because the classifier trained on the training samples was already well-calibrated for the familiar validation samples.  We also found calibration to improve the Bayesian method~\cite{kendall2017uncertainties}.  Calibration has little effect for distillation and G-distillation, likely because distillation's fitting to soft labels makes it less confident.  For those methods, we used calibration only when $T>=1$, as setting $T<1$ always made classifiers more over-confident. In Table~\ref{tab:bigtable}, ``T-scaling'' refers to the T-scaled baseline, and T-scaling is used for all other non-baseline methods as well.

Given the benefits of T-scaling, we expected that novelty-weighted scaling, in which samples predicted to be unfamiliar have a greater temperature (reducing confidence more), would further improve results.  However, we found the novelty weight $T_1$ was usually set to zero in validation, and, in any case, the novelty-weighted scaling performed similarly to T-scaling.  The problem could be that the validation set does not have enough novelty to determine the correct weights.  If we ``cheat'' and use samples drawn from the unfamiliar distribution to set the two weights $T_0$ and $T_1$, the method performs quite well.  For example, when tuning parameters on a mix of familiar and unfamiliar samples for Gender recognition, novelty-weighted scaling performed best with 0.297 NLL compared to 0.328 for T-scaling and 0.313 for ensemble of calibrated classifiers that are tuned on the same data. 

\begin{figure*}[t]
  \centering
    \subfigure[Gender recognition]{
        \includegraphics[width=0.305\textwidth]{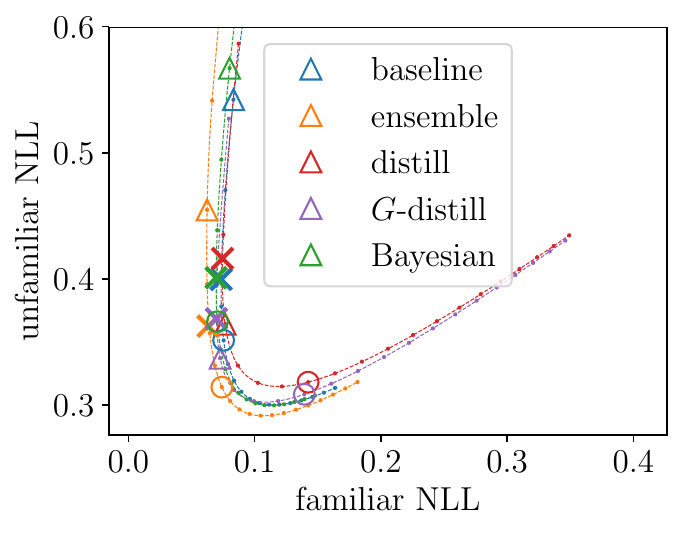}
        \label{fig:curves_gender}
    }
    \subfigure[Cats vs. Dogs]{
        \includegraphics[width=0.315\textwidth]{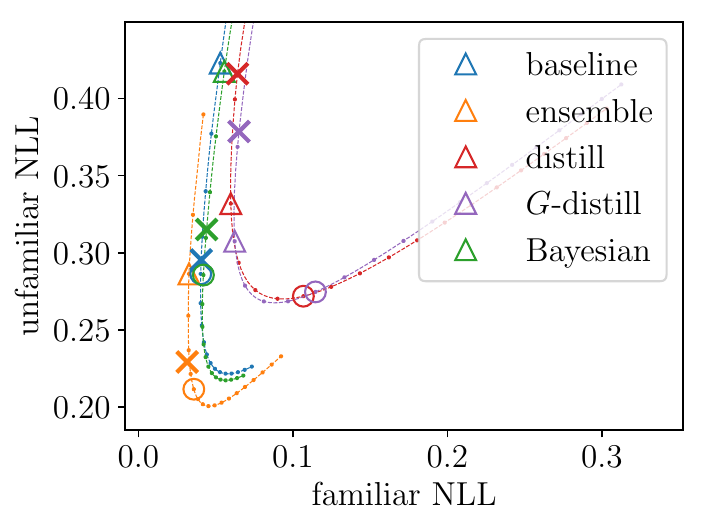}
        \label{fig:curves_cat_dog}
    }
    \subfigure[Animal Categorization]{
        \includegraphics[width=0.31\textwidth]{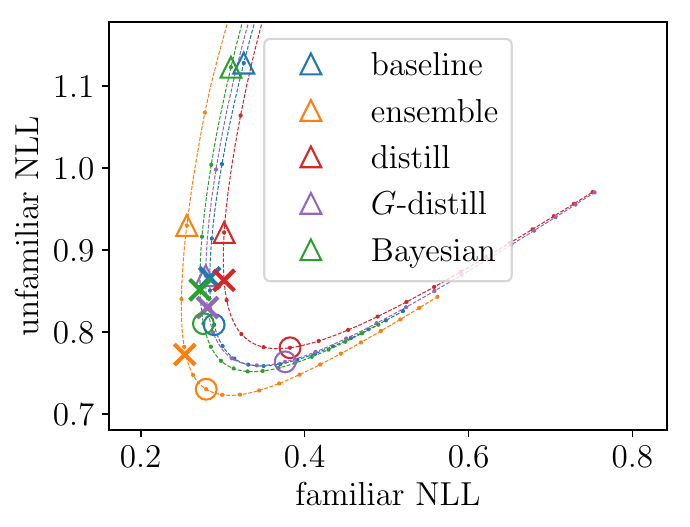}
        \label{fig:curves_animals}
    }
  \caption{Familiar and unfamiliar NLL error while varying the $T$ calibration parameter. Triangles mark the uncalibrated models; `X' marks models calibrated on the validation set.  Circles mark $T=2$, with each rightward dot increasing by 0.25.  Without calibration, classifiers are often overconfident even for familiar samples, so calibration reduces confidence to improve NLL for familiar and unfamiliar.  Ensembles dominate the other methods, always achieving lower NLL for some $T$. }
  \label{fig:curves}
  \vspace{-0.1in}
\end{figure*}

In Figure~\ref{fig:curves}, we plot calibration curves of single networks, ensembles, distillation, G-distillation, and the Bayesian method with varying $T$.  These curves allow us to peek at the best possible performance, if we were able to tune calibration parameters on unfamiliar and familiar test data.  These curves allow a clearer view of which methods perform best.  They also show that calibration on the familiar samples (`X' marks) leads to lower $T$ values than is optimal for the unfamiliar samples.  Generally, increasing $T$ further would reduce likelihood error for unfamiliar samples without much adverse impact on likelihood error for familiar samples.  On the object presence task (curve not shown), all models are well-calibrated (without T-scaling) for both unfamiliar and familiar categories.

\textbf{Comparison of methods: } Finally, considering Table~\ref{tab:bigtable}, we see that the ensemble of T-scaled models dominates, consistently achieving the lowest label error, calibration error, NLL, and Brier error.  The downside of the ensemble is higher training and inference computational cost, 10x in our case since we test with an ensemble of 10 classifiers.  Distillation and G-distillation offered hope of preserving some of the gains of ensembles without the cost, and we expected the performance of G-distillation at least to fall between T-scaling and the ensemble. However, while G-distillation, which uses unsupervised samples to better mimic ensemble behavior in the broader feature domain, slightly outperforms distillation, neither method  consistently outperforms T-scaling --- no pain, no gain.  

The method of Kendall et al.~\cite{kendall2017uncertainties}, which we call ``Bayesian'', performs second best to the ensemble, with small reductions in label error and comparable calibration improvements to all methods except ensemble. The Bayesian method also requires generating multiple predictions via MC-Dropout at test time, so also incurs significant additional computational cost.   


The supplemental material compares prediction entropy to label cross-entropy (NLL), showing that calibration eliminates overconfidence for familiar samples but calibrated ensembles further reduce overconfidence on unfamiliar samples by increasing prediction uncertainty and improving accuracy. 

\subsection{Findings}

We summarize our findings:
\begin{itemize}
    \itemsep0em
    \item Unfamiliar samples lead to much higher calibration error and label error, which can make their behavior unreliable in applications for which inputs are sampled differently in training and deployment.  
    \item T-scaling is effective in reducing likelihood and calibration error on familiar and unfamiliar samples.
    \item The simple ensemble, when applied to T-scaled models, is the best method overall, reducing all types of error for both unfamiliar and familiar samples.  The method of Kendall et al.~\cite{kendall2017uncertainties} is the only other tested method to consistently reduce labeling error.
    \item T-scaling, distillation, and G-distillation all perform much better than the baseline.
\end{itemize}

Our recommendation: developers of any application that relies on prediction confidences (e.g. deciding whether to return a label, or to sound an alarm) should calibrate their models or, better yet, use calibrated ensembles.  Ensembles achieve higher accuracy and better calibration, but at additional computational expense.  We suspect that ensembles of shallower networks may outperform single deeper networks with similar computation costs, though we leave confirmation to future work. Tuning calibration on a validation set that is i.i.d with training leads to overestimates of confidence for unfamiliar samples, so to minimize likelihood error for both unfamiliar and familiar samples, it may be best to obtain a small differently-sampled validation set.



\section{Conclusion}
We show that modern deep network classifiers are prone to overconfident errors, especially for unfamiliar but valid samples. We show that ensembles of T-scaled models are best able to reduce all kinds of prediction error.  Our work is complementary to recent works on calibration of i.i.d. data (e.g., Guo et al.~\cite{guo2017calibration}) and artificially distorted data~\cite{ovadia2019}.  More work is needed to improve prediction reliability with a single model in the unfamiliar setting and to consolidate learnings from the multiple recent studies of calibration and generalization. Data augmentation and representation learning are other important ways to improve generalization, and it would be interesting to evaluate their effect on prediction for both familiar and unfamiliar samples.

\vspace{0.05in}
{\small \noindent\textbf{Acknowledgments}: 
This work is supported in part by NSF Award IIS-1421521 and by Office of Naval Research grant ONR MURI N00014-16-1-2007. }

{\small
\bibliographystyle{ieee_fullname}
\bibliography{main}
}
\storecounter{figure}{figurecounterstore}

\appendix
\section*{Appendix}
In Sec.~\ref{sec:entropy}, we compare the entropy and cross-entropy (NLL) of three approaches to analyze overconfidence.

In Sec.~\ref{sec:toy}, we show experimental results on a simple dataset to illustrate why ensembles perform well for unfamiliar samples and how use of unsupervised samples by G-distill can lead it to mimic the performance of the ensemble (at least in the ideal case where unsupervised samples cover a superset of the unfamiliar samples).  

In Sec.~\ref{sec:complete}, we show the complete table of results, mainly to simplify comparisons by any later works. Note that in the supplemental material methods without ``T-scaling'' in the name do not use calibration.  In the main table of the paper, for brevity, only results with calibration are shown except where noted.  So ``Ensemble'' in the main paper is ``Ensemble of T-scaled models'' here.  

In Sec.~\ref{sec:densenet}, we show results on one of the tasks with DenseNet-161, supporting the same conclusions as we found based on experiments with ResNet-18.  We leave a more complete exploration of depth and architecture of network to future work.

\begin{figure*}[t]
  \centering
    \includegraphics[width=0.84\textwidth]{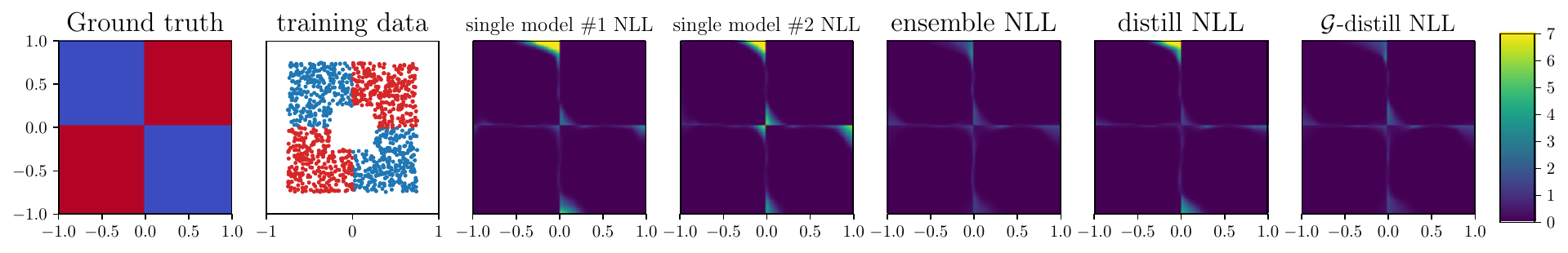}
    \includegraphics[width=0.15\textwidth]{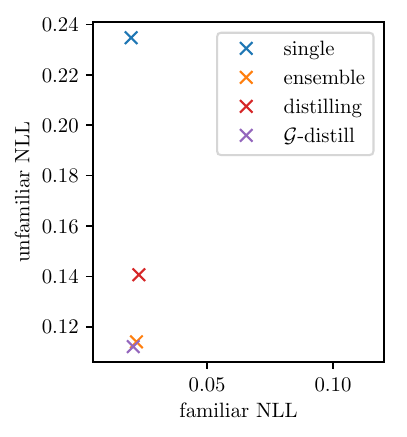}
    \includegraphics[width=0.84\textwidth]{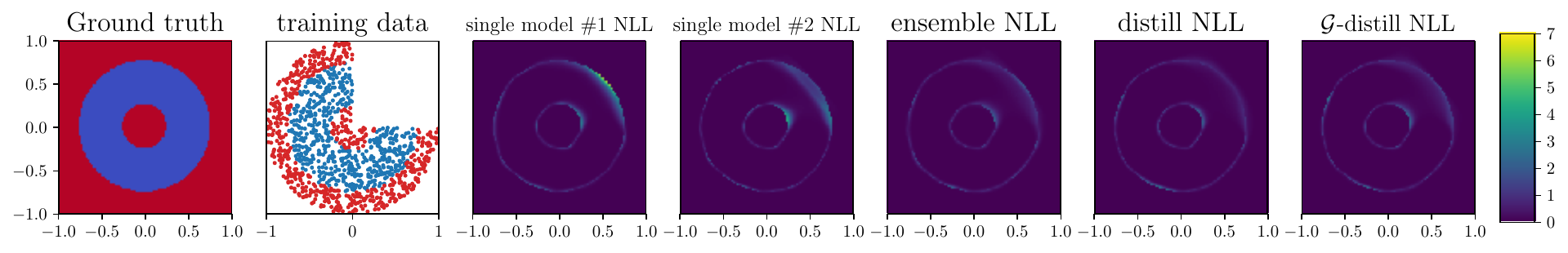}
    \includegraphics[width=0.15\textwidth]{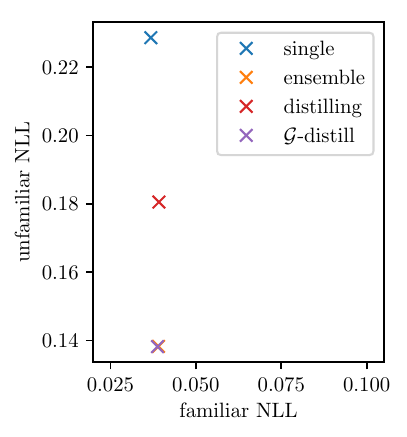}
    \includegraphics[width=0.84\textwidth]{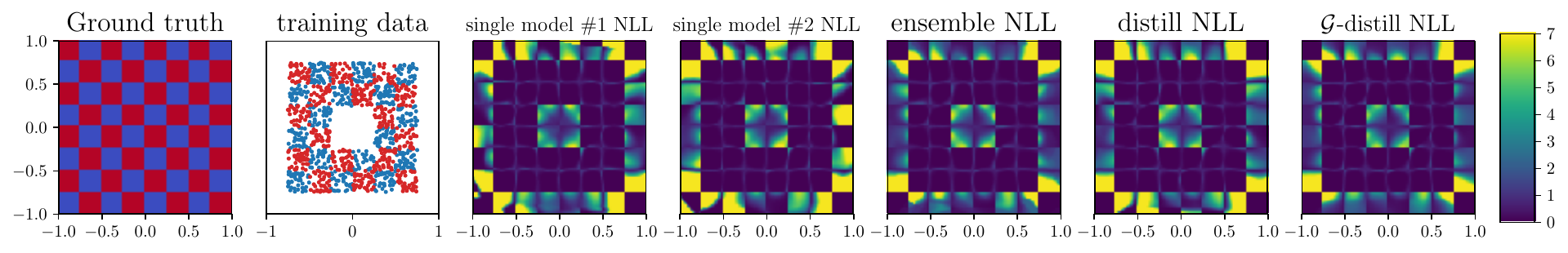}
    \includegraphics[width=0.15\textwidth]{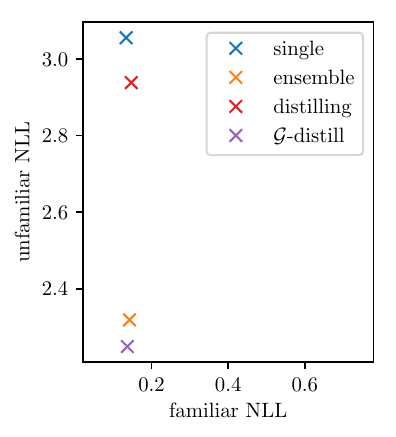}
    \includegraphics[width=0.84\textwidth]{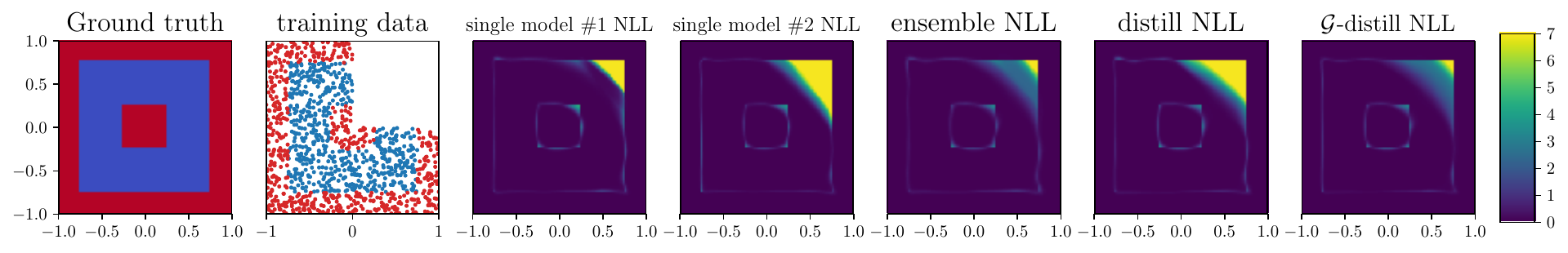}
    \includegraphics[width=0.15\textwidth]{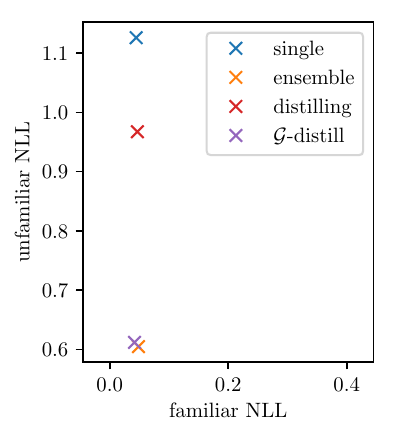}
    \caption{\textbf{Illustration on toy datasets.} Ground truth for each class are shown in red and blue. ``Familiar'' data is sampled from a portion of the 2D feature space. Negative log likelihood (NLL) errors are shown for two single models, ensembles, and two distillation methods.  On right, average NLL for familiar and unfamiliar samples are shown.  Different single models can make mistakes in different areas, while ensembles average out these differences.  Distillation, when based only on familiar samples, fails to mimic the ensemble's behavior in the unfamiliar areas.  G-distillation, which incorporates unsupervised unfamiliar samples, performs similarly to the ensemble but does not require multiple models at test time.  In experiments on real data, however, (see main paper), G-distill underperforms the ensemble, likely because it is not possible to densely sample the unfamiliar space in practice.  Figure best read in color.}
    \label{exfig:toy}
\end{figure*}

\section{Entropy vs. NLL/Cross-Entropy}
\label{sec:entropy}
We thank one of the reviewers for suggesting analysis of prediction entropy, which we include in  Table~\ref{tab:entropy}.  Prediction entropy measures the uncertainty of classification and is maximized if the classifier outputs uniform probabilities for each class.  NLL, equivalent to cross-entropy when using hard labels, measures the uncertainty in the correct label.  When entropy is lower than cross-entropy (i.e. more confident than confidently correct), the classifier is overconfident.



\begin{table}[ht!]
\small
\setlength{\tabcolsep}{5pt}
\begin{tabular}{lcccccc}
                & \multicolumn{2}{c}{\textbf{NLL}} & \multicolumn{2}{c}{\textbf{Entropy}} & \multicolumn{2}{c}{\textbf{NLL-Ent.}} \\
\textbf{Gender} & \textbf{fam.}  & \textbf{unf.}  & \textbf{fam.}    & \textbf{novel}    & \textbf{fam.}      & \textbf{unf.}      \\
Single                                                       & 0.083     & 0.542     & 0.036     & 0.089     & 0.047     & 0.453     \\
Sin. T-scale                                                 & 0.073     & 0.400     & 0.069     & 0.139     & 0.005     & 0.261     \\
Ens. T-scale & 0.063     & 0.363     & 0.079     & 0.158     & -0.016    & 0.205     \\
                                                             &           &           &           &           &           &           \\
\textbf{Cat vs. Dog}                                         & \textbf{} & \textbf{} & \textbf{} & \textbf{} & \textbf{} & \textbf{} \\
Single                                                       & 0.053     & 0.423     & 0.014     & 0.043     & 0.039     & 0.380     \\
Sin. T-scale                                                 & 0.041     & 0.295     & 0.033     & 0.090     & 0.007     & 0.206     \\
Ens. T-scale & 0.032     & 0.229     & 0.039     & 0.113     & -0.007    & 0.116     \\
                                                             &           &           &           &           &           &           \\
\textbf{Animals}                                             & \textbf{} & \textbf{} & \textbf{} & \textbf{} & \textbf{} & \textbf{} \\
Single                                                       & 0.326     & 1.128     & 0.146     & 0.263     & 0.180     & 0.864     \\
Sin. T-scale                                                 & 0.284     & 0.866     & 0.282     & 0.451     & 0.002     & 0.415     \\
Ens. T-scale & 0.254     & 0.772     & 0.311     & 0.504     & -0.057    & 0.269     \\
\end{tabular}
\hrule
\vspace{0.02in}
\caption{
\normalsize
\label{tab:entropy}
When prediction entropy is lower than NLL (cross-entropy), the classifier is overconfident, i.e. more confident than confidently correct. We see, e.g., that single-model calibration eliminates overconfidence for familiar examples, but the calibrated ensemble achieves much further reduction in overconfidence for unfamiliar examples by increasing uncertainty and improving accuracy (lower NLL and label error).
\vspace{-0.05in}
}
\end{table}



\section{Toy Experiment}
\label{sec:toy}

Figure~\ref{exfig:toy} shows results of single models, ensembles, and distillation models on simple datasets with two dimensional features.  We take 1200 samples for both train and validation. The test set is densely sampled.  For these experimetns, we use a 3-hidden-layer network, both layers with 1024 hidden units and Glorot initialization similar to popular deep networks, to avoid bad local minima when layer widths are too small~\cite{choromanska2015loss}. Batchnorm~\cite{ioffe2015batchnorm} and then dropout~\cite{srivastava2014dropout} are applied after ReLU. The same hyperparameter tuning, initialization, and training procedures are used as described in the main paper.

\section{Complete Results Table}
\label{sec:complete}
Table~\ref{tab:bigtable} shows the complete table of absolute errors for all methods tested across all datasets.  In the main paper, a subset of methods is shown due to space constraints (and to save the reader from being overwhelmed), with performance relative to baseline (single model) shown. This table is provided for completeness and to facilitate comparison by other methods.

\begin{table*}[t]
\setlength{\tabcolsep}{3.5pt}
\small
\begin{tabular}{lllllllllll}
 & \multicolumn{2}{c}{\textbf{NLL}} & \multicolumn{2}{c}{\textbf{Brier}} & \multicolumn{2}{c}{\textbf{Label Error}} & \multicolumn{2}{c}{\textbf{ECE}} & \multicolumn{2}{c}{\textbf{E99}} \\
 \textbf{Gender} & \textbf{familiar} & \textbf{unfam.} & \textbf{familiar} & \textbf{unfam.} & \textbf{familiar} & \textbf{unfam.} & \textbf{familiar} & \textbf{unfam.} & \textbf{familiar} & \textbf{unfam.} \\
Single Model & 0.08324 & 0.54208 & 0.14663 & 0.35199 & 0.02772 & \textbf{0.14682} & 0.01348 & 0.10902 & 0.00470 & 0.06021 \\
Single + T-scaling & 0.07348 & 0.39971 & 0.14332 & \textbf{0.33715} & 0.02772 & \textbf{0.14682} & 0.00361 & 0.08737 & 0.00192 & 0.02324 \\
Ensemble & \textbf{0.06233} & 0.45471 & \textbf{0.13077} & 0.34312 & \textbf{0.02195} & \textbf{0.14714} & \textbf{0.00272} & 0.09341 & 0.00171 & 0.03856 \\
Ensemble of T-scaled models & \textbf{0.06312} & 0.36266 & \textbf{0.13153} & \textbf{0.33246} & \textbf{0.02170} & \textbf{0.14714} & 0.00856 & \textbf{0.07723} & \textbf{0.00131} & 0.01003 \\
Distill & 0.07661 & 0.36426 & 0.14283 & 0.33963 & 0.02690 & 0.15641 & 0.00797 & 0.08596 & \textbf{0.00141} & \textbf{0.00244} \\
Distill + T-scaling & 0.07457 & 0.41629 & 0.14304 & 0.34873 & 0.02690 & 0.15641 & 0.00532 & 0.09974 & 0.00230 & 0.01456 \\
G-distill & 0.07268 & \textbf{0.33729} & 0.13885 & \textbf{0.33216} & 0.02519 & 0.15346 & 0.00928 & \textbf{0.07535} & \textbf{0.00117} & \textbf{0.00121} \\
G-distill + T-scaling & 0.06972 & 0.36859 & 0.13853 & 0.33913 & 0.02519 & 0.15346 & 0.00416 & 0.08600 & 0.00174 & 0.00424 \\
Novelty scaling & 0.07348 & 0.39971 & 0.14332 & \textbf{0.33715} & 0.02772 & \textbf{0.14682} & 0.00361 & 0.08737 & 0.00192 & 0.02324 \\
Bayesian & 0.08005 & 0.56668 & 0.14216 & 0.35391 & 0.02570 & \textbf{0.14709} & 0.01249 & 0.11104 & 0.00504 & 0.06768 \\
Bayesian + T-scaling & 0.06955 & 0.40056 & 0.13907 & 0.33765 & 0.02585 & \textbf{0.14797} & \textbf{0.00315} & 0.08884 & 0.00165 & 0.02300 \\
 &  &  &  &  &  &  &  &  &  &  \\
\textbf{Cat vs. Dog}  &  &  &  &  &  &  &  &  &  &  \\
Single Model & 0.05296 & 0.42285 & 0.11158 & 0.29026 & 0.01555 & 0.09518 & 0.00976 & 0.07777 & 0.00394 & 0.05251 \\
Single + T-scaling & 0.04059 & 0.29537 & 0.10686 & 0.27653 & 0.01555 & 0.09518 & 0.00356 & 0.05953 & 0.00074 & 0.02212 \\
Ensemble & \textbf{0.03271} & 0.28633 & \textbf{0.09343} & 0.26247 & \textbf{0.01180} & \textbf{0.08756} & 0.00248 & 0.05493 & 0.00074 & 0.02396 \\
Ensemble of T-scaled models & \textbf{0.03154} & \textbf{0.22931} & \textbf{0.09252} & \textbf{0.25532} & \textbf{0.01215} & \textbf{0.08756} & \textbf{0.00202} & \textbf{0.04222} & \textbf{0.00040} & \textbf{0.01313} \\
Distill & 0.05975 & 0.33184 & 0.12161 & 0.28798 & 0.01836 & 0.09937 & 0.00438 & 0.05785 & 0.00193 & 0.03610 \\
Distill + T-scaling & 0.06411 & 0.41595 & 0.12309 & 0.29453 & 0.01836 & 0.09937 & 0.00860 & 0.07354 & 0.00490 & 0.05314 \\
G-distill & 0.06232 & 0.30747 & 0.12732 & 0.28745 & 0.02065 & 0.10254 & 0.00577 & 0.05390 & 0.00123 & 0.02163 \\
G-distill + T-scaling & 0.06509 & 0.37850 & 0.12892 & 0.29445 & 0.02065 & 0.10254 & 0.00857 & 0.07182 & 0.00314 & 0.04420 \\
Novelty scaling & 0.04024 & 0.29713 & 0.10635 & 0.27432 & 0.01555 & 0.09518 & \textbf{0.00255} & 0.05662 & 0.00081 & 0.02396 \\
Bayesian & 0.05551 & 0.41758 & 0.11221 & 0.28485 & 0.01541 & 0.09264 & 0.00986 & 0.07454 & 0.00444 & 0.05306 \\
Bayesian + T-scaling & 0.04381 & 0.31260 & 0.10826 & 0.27497 & 0.01558 & 0.09264 & 0.00563 & 0.06152 & 0.00173 & 0.02825 \\
 &  &  &  &  &  &  &  &  &  &  \\
\textbf{Animals}  &  &  &  &  &  &  &  &  &  &  \\
Single Model & 0.32575 & 1.12785 & 0.19922 & 0.34062 & 0.10375 & 0.29056 & 0.04807 & 0.18714 & 0.01339 & 0.08701 \\
Single + T-scaling & 0.28425 & 0.86575 & 0.19386 & 0.32398 & 0.10375 & 0.29056 & \textbf{0.01208} & 0.11751 & \textbf{0.00219} & 0.02567 \\
Ensemble & 0.25623 & 0.92980 & \textbf{0.18108} & 0.32221 & 0.09437 & \textbf{0.27563} & 0.02236 & 0.13766 & 0.00521 & 0.04509 \\
Ensemble of T-scaled models & \textbf{0.25377} & \textbf{0.77222} & \textbf{0.18149} & \textbf{0.31193} & \textbf{0.09250} & \textbf{0.27438} & 0.02408 & \textbf{0.07979} & \textbf{0.00174} & \textbf{0.01329} \\
Distill & 0.30180 & 0.92112 & 0.19639 & 0.32732 & 0.10450 & 0.29000 & \textbf{0.01329} & 0.12952 & 0.00650 & 0.04228 \\
Distill + T-scaling & 0.30167 & 0.86280 & 0.19657 & 0.32303 & 0.10450 & 0.29000 & 0.01646 & 0.10353 & 0.00481 & 0.03457 \\
G-distill & 0.27929 & 0.86841 & 0.18950 & 0.32109 & 0.09644 & 0.28444 & 0.01489 & 0.11629 & 0.00651 & 0.03399 \\
G-distill + T-scaling & 0.28152 & 0.82950 & 0.19011 & 0.31806 & 0.09644 & 0.28444 & 0.02105 & 0.09629 & 0.00302 & 0.03354 \\
Novelty scaling & 0.28425 & 0.86575 & 0.19386 & 0.32398 & 0.10375 & 0.29056 & \textbf{0.01208} & 0.11751 & \textbf{0.00219} & 0.02567 \\
Bayesian & 0.30986 & 1.12297 & 0.19370 & 0.33759 & 0.09906 & 0.28694 & 0.04440 & 0.18238 & 0.01439 & 0.09471 \\
Bayesian + T-scaling & 0.27239 & 0.86154 & 0.18905 & 0.32226 & 0.09863 & 0.28694 & \textbf{0.01437} & 0.11515 & \textbf{0.00290} & 0.03181 \\
 &  &  &  &  &  &  &  &  &  &  \\
\textbf{Objects}  &  &  &  &  &  &  &  &  &  &  \\
Single Model & 0.08597 & 0.12815 & 0.15392 & 0.18553 & 0.19494 & 0.45523 & 0.00475 & 0.01021 & 0.00222 & 0.00546 \\
Single + T-scaling & 0.08589 & 0.12780 & 0.15388 & 0.18550 & 0.19494 & 0.45523 & 0.00466 & 0.01003 & 0.00207 & 0.00519 \\
Ensemble & \textbf{0.08222} & 0.12292 & \textbf{0.15063} & 0.18207 & \textbf{0.18298} & \textbf{0.44095} & \textbf{0.00435} & 0.00950 & 0.00179 & 0.00459 \\
Ensemble of T-scaled models & \textbf{0.08227} & 0.12274 & \textbf{0.15063} & 0.18207 & \textbf{0.18299} & \textbf{0.44095} & 0.00459 & 0.00953 & 0.00171 & 0.00437 \\
Distill & 0.08658 & \textbf{0.12165} & 0.15437 & \textbf{0.18166} & 0.19308 & 0.45322 & 0.00624 & \textbf{0.00918} & \textbf{0.00122} & 0.00325 \\
Distill + T-scaling & 0.08583 & 0.12218 & 0.15421 & \textbf{0.18160} & 0.19308 & 0.45322 & 0.00450 & \textbf{0.00903} & 0.00191 & 0.00456 \\
G-distill & 0.08736 & 0.12196 & 0.15527 & 0.18188 & 0.19822 & 0.45861 & 0.00670 & 0.00951 & \textbf{0.00119} & \textbf{0.00315} \\
G-distill + T-scaling & 0.08661 & 0.12229 & 0.15511 & \textbf{0.18180} & 0.19822 & 0.45861 & 0.00485 & \textbf{0.00905} & 0.00187 & 0.00441 \\
Novelty scaling & 0.08582 & 0.12809 & 0.15385 & 0.18561 & 0.19452 & 0.45573 & 0.00460 & 0.01007 & 0.00205 & 0.00516 \\
Bayesian & 0.08597 & 0.12884 & 0.15359 & 0.18577 & 0.19440 & 0.45674 & 0.00474 & 0.01046 & 0.00254 & 0.00581 \\
Bayesian + T-scaling & 0.08580 & 0.12789 & 0.15356 & 0.18569 & 0.19444 & 0.45686 & 0.00460 & 0.01008 & 0.00211 & 0.00504 \\
\bottomrule
\end{tabular}
\vspace{0.02in}
\caption{
\normalsize
Errors of all tested methods across all datasets.  Bold numbers are best or not significantly different than the best.  
\label{tab:bigtable}
}
\end{table*}%

\section{Results on DenseNet-161}
\label{sec:densenet}
We also ran with all models on the Gender task using the DenseNet-161 architecture, as shown in Table~\ref{tab:densenet}.  In this case Dropout was used for all layers of the network for ``Bayesian''.  Ensemble of T-scaled Networks is still the clear leader for this architecture.

\begin{table*}[t]
\small
\begin{tabular}{lllllllllll}
 & \multicolumn{2}{c}{\textbf{NLL}} & \multicolumn{2}{c}{\textbf{Brier}} & \multicolumn{2}{c}{\textbf{Label Error}} & \multicolumn{2}{c}{\textbf{ECE}} & \multicolumn{2}{c}{\textbf{E99}} \\
\textbf{Gender} & \textbf{familiar} & \textbf{unfam.} & \textbf{familiar} & \textbf{unfam.} & \textbf{familiar} & \textbf{unfam.} & \textbf{familiar} & \textbf{unfam.} & \textbf{familiar} & \textbf{unfam.} \\
Single Model & 0.0769 & 0.5608 & 0.1332 & 0.3499 & 0.0219 & 0.1430 & 0.0132 & 0.1139 & 0.0063 & 0.0658 \\
Single + T-scaling & 0.0611 & 0.3553 & 0.1291 & 0.3262 & 0.0219 & 0.1430 & \textbf{0.0024} & 0.0850 & 0.0015 & 0.0131 \\
Ensemble & \textbf{0.0513} & 0.4103 & \textbf{0.1163} & 0.3281 & \textbf{0.0180} & \textbf{0.1342} & \textbf{0.0031} & 0.0861 & 0.0022 & 0.0348 \\
Ensemble of T-scaled models & \textbf{0.0507} & \textbf{0.2995} & \textbf{0.1165} & \textbf{0.3116} & \textbf{0.0185} & \textbf{0.1338} & 0.0070 & \textbf{0.0653} & \textbf{0.0003} & \textbf{0.0023} \\
Distill & 0.0706 & 0.4117 & 0.1321 & 0.3347 & 0.0211 & \textbf{0.1352} & 0.0081 & 0.0951 & 0.0039 & 0.0245 \\
Distill + T-scaling & 0.0694 & 0.3947 & 0.1317 & 0.3326 & 0.0211 & \textbf{0.1352} & 0.0067 & 0.0920 & 0.0034 & 0.0186 \\
G-distill & 0.0645 & 0.3559 & 0.1265 & 0.3250 & 0.0196 & 0.1391 & 0.0049 & 0.0815 & 0.0030 & 0.0138 \\
G-distill + T-scaling & 0.0641 & 0.3477 & 0.1263 & 0.3235 & 0.0196 & 0.1391 & 0.0041 & 0.0791 & 0.0026 & 0.0118 \\
Novelty scaling & 0.0611 & 0.3553 & 0.1291 & 0.3262 & 0.0219 & 0.1430 & \textbf{0.0024} & 0.0850 & 0.0015 & 0.0131 \\
Bayesian & 0.0795 & 0.5930 & 0.1341 & 0.3512 & 0.0218 & 0.1416 & 0.0139 & 0.1155 & 0.0070 & 0.0738 \\
Bayesian + T-scaling & 0.0617 & 0.3934 & 0.1295 & 0.3324 & 0.0217 & 0.1412 & 0.0050 & 0.0932 & 0.0018 & 0.0206 \\
\bottomrule
\end{tabular}
\vspace{0.02in}
\caption{
\normalsize
Performance for DenseNet-161 classifier.  Best and within significance range of best is in bold.  
\label{tab:densenet}
}
\end{table*}

\end{document}